\def\year{2022}\relax
\documentclass[letterpaper]{article} 
\usepackage{aaai22}  
\usepackage{times}  
\usepackage{helvet}  
\usepackage{courier}  
\usepackage[hyphens]{url}  
\usepackage{graphicx} 
\usepackage{natbib}  
\usepackage{caption} 
\DeclareCaptionStyle{ruled}{labelfont=normalfont,labelsep=colon,strut=off} 
\usepackage{algorithm}
\usepackage{algorithmic}

\usepackage{newfloat}
\usepackage{listings}
\floatstyle{ruled}
\newfloat{listing}{tb}{lst}{}
\floatname{listing}{Listing}
\usepackage{researchpack}
\usepackage{xcolor}
\usepackage{xspace}

\newcommand{\av}[1]{\textcolor{green}{\textbf{[AV: #1]}}}

\newcommand{\acronym}{Conditional Randomized Interactive Skeptical Probabilistic circuits}
\newcommand{\method}{\textsc{crisp}\xspace}
\newcommand{\methods}{\textsc{crisp}s\xspace}
\newcommand{\Renyi}{R\'enyi\xspace}

\newcommand{\val}{\ensuremath{\mathsf{val}}}

\newcommand{\ch}{\ensuremath{\mathsf{in}}}
\newcommand{\scope}{\ensuremath{{\mathsf{scope}}}}

\newcommand{\defeq}{\ensuremath{:=}}
\newcommand{\noisyy}{\ensuremath{\widetilde{y}}\xspace}
\newcommand{\noisyvy}{\ensuremath{\widetilde{\vy}}\xspace}

\newcommand{\noisyvY}{\ensuremath{\widetilde{\vY}}\xspace}

\newcommand{\predvy}{\ensuremath{\widehat{\vy}}\xspace}

\newcommand{\BK}{\ensuremath{\mathsf{K}}\xspace}

\title{Efficient and Reliable Probabilistic Interactive Learning with Structured Outputs}
\author {
    Stefano Teso,\textsuperscript{\rm 1}
    Antonio Vergari\textsuperscript{\rm 2}
}
\affiliations {
    \textsuperscript{\rm 1} University of Trento, Italy\\
    \textsuperscript{\rm 2} University of Edinburgh, UK\\
    stefano.teso@unitn.it, avergari@ed.ac.uk
}

\begin{document}
\maketitle

\begin{abstract}
    In this position paper, we study interactive learning for structured output spaces, with a focus on \emph{active} learning, in which labels are unknown and must be acquired, and on \emph{skeptical} learning, in which the labels are noisy and may need relabeling.
    These scenarios require expressive models that guarantee reliable and efficient computation of probabilistic quantities to measure uncertainty.
    We identify conditions under which a class of probabilistic models---which we denote \methods---meet all of these conditions, thus delivering tractable computation of the above quantities while preserving expressiveness.
    Building on prior work on tractable probabilistic circuits, we illustrate how \methods enable robust and efficient active and skeptical learning in large structured output spaces.
\end{abstract}

\section{Introduction}

All probabilistic models learned from data are \textit{wrong}, in the sense that they can be arbitrarily far from the data distribution.  Models that can \textit{reliably} quantify uncertainties about the distribution they encode, however, are \textit{useful}.  
Reliable inference with guarantees indeed is the key for enabling learning ``in the wild'', especially when dealing with (unreliable) human annotators.
This becomes evident in the context of \textit{interactive learning} with large structured output spaces~\cite{tsochantaridis2004support}.
In the wild, misspecification errors can quickly pile up as the learner receives noisy supervision.
In a classic \textit{active learning} scenario, the predictive accuracy of the model depends chiefly on its ability to (reliably) know when it does not know ``enough'' about a sample, in which case it should opt to obtain the label from a user~\citep{settles2012active,gal2017deep}.
As providing all labels for a single sample in a structured-output prediction (SOP) task might be infeasible or highly expensive in practice, an efficient probabilistic model should be able to identify informative \textit{subset of labels} to be annotated by the user.
Furthermore, we can improve uncertainty estimates even more if our probabilistic model is able to question if the user-provided labels are correct, a setting recently proposed under the name of \textit{skeptical learning}~\citep{zeni2019fixing,bontempelli2020learning}.

In this position paper, we investigate when and how a probabilistic model can satisfy the aforementioned desiderata in these settings.
We define \methods (\acronym), a class of tractable probabilistic models that support reliable and efficient active and skeptical learning with full and partial labels.  \methods exploit recent advancements in the literature of probabilistic circuits~\citep{choi2020pc,vergari2021compositional}, deep computational graphs encoding complex probability distributions that support exact computation of 
the uncertainties needed for incremental learning~\cite{vergari2021compositional}.

\methods offer several benefits.
First, they leverage the expressiveness of deep neural classifiers by implementing a conditional probabilistic circuit~\citep{shao2020conditional} that is able to encode intricate dependencies over the labels and deal with sub-symbolic inputs.
As such, they can act as a \textit{drop-in replacement for the ubiquitous, but generally intractable, softmax/sigmoid last layer} in deep neural networks.
Second, by ensuring that \methods satisfy certain structural properties, we can \textit{always guarantee that these quantities can be computed exactly and in time linear} with respect to the size of the computational graph that encodes the \method model.
To circumvent the need to learn the structure of the computational graph, we propose to leverage a randomized construction approach~\citep{peharz2020random,shih2020probabilistic} that guarantees the structural properties we need.

To the best of our knowledge, this is the first paper to investigate the benefits of tractable probabilistic inference for reliable interactive learning in the wild  and how to combine it in a principled way with the intractable, unreliable but largely used deep learning paradigm.

\section{Probabilistic Interactive Learning:\\
Problem Statement and Inference Tasks}

We consider structured prediction tasks in which a machine learns to associate instances $\vx \in \bbR^d$ to $c$ \emph{interdependent} concepts, identified by labels $\vy \in \{0, 1\}^c$.
Dependencies among labels are sometimes realized as \textit{correlations}, as is the case in multi-label classification~\citep{dembczynski2012label}.
Other times, they occur as \textit{hard constraints} and are expressed as logical formulas encoding, e.g., a hierarchy over the labels~\cite{giunchiglia2020coherent}.
We focus on two challenging sequential learning tasks, \emph{active learning} and \emph{skeptical learning}, and illustrate which uncertainty-based queries are necessary to solve them.

%
%
\paragraph{Active learning.}  In incremental active learning~\cite{hoi2021online}, {aka} \textit{selective sampling},\footnote{Despite our focus on sequential tasks, most of our results do carry over to pool-based active learning~\cite{settles2012active}.} the machine receives a sequence of instances $\vx_1, \vx_2, \ldots,$ and for each one of them it has to decide whether to query a human supervisor---at a non-negligible cost---to obtain the corresponding \textit{complete} ground truth label ${\vy}_t$.
This decision depends crucially on how confident the model is in its own prediction
$
    \hat{\vy}_t = \argmax_{\vy} p_{\theta}(\vy\mid\vx_t)
$,
where $p_{\theta}$ denotes the model's conditional distribution.
Ultimately, the model requests a ground-truth label $\vy_t$ if it holds that:
\[
    U_\theta(\vx_t) \geq T
\]
where $U_\theta(\vx)$ quantifies how uncertain the model is about the label of instance $\vx$, and $T$ is a custom threshold.

A popular way of modelling $U_\theta(\vx)$ is to use the \textit{Shannon entropy} of $p_{\theta}$, denoted as $H_\theta(\vx)$, which is defined as:
\[
    \textstyle
    H_\theta(\vx) \defeq - \sum_{\vy \in \{0, 1\}^c} p_\theta(\vy\mid\vx) \log p_\theta(\vy\mid\vx).
    \label{eq:entropy}
\]
Another option is to compute the probabilistic \textit{margin}, defined as the distance of the mode of the distribution from absolute certainty:
\[
    \textstyle
    M_\theta(\vx) \defeq 1 - \max_{\vy} \ p_\theta(\vy\mid\vx).
    \label{eq:margin}
\]
Acquiring the label of uncertain instances is guaranteed to reduce the model's total uncertainty.
Unfortunately, evaluating Eqs.~\ref{eq:entropy} and \ref{eq:margin} is NP-hard for general probabilistic models~\cite{vergari2021compositional,darwiche2002knowledge}. 

Popular solutions to this problem include using a sigmoid final classification layer in deep neural networks, which renders marginalization and maximization over the labels straightforward.  This, however, amounts to assuming all labels to be independent given the features $\vx$.
Clearly, this simplifying assumption discards precious information about the label dependencies, which is crucial in several SOP tasks such as multi-label classification~\citep{ghamrawi2005collective}.
Furthermore, the sigmoid layer solution has been consistently reported to deliver poorly calibrated uncertainties in deep learning models~\citep{gal2017deep,guo2017calibration}.
Other common workarounds employing low-order approximate models~\citep{behpour2019active} fall prey of the same issues.

Solutions based on expected gradient length (EGL) define uncertainty in terms of the (squared norm of the) expected change in model parameters $\theta$ after receiving the label of a query~\cite{settles2007multiple}.  The EGL can be computed rather cheaply and exactly for differentiable models, but it tends to perform sub-optimally.  This is partly because the norm of the gradient depends on the encoding of the parameters.  A standard solution is to compute the \textit{natural gradient} instead, which however involves evaluating the Fisher information matrix~\cite{kunstner2019limitations}.  This is computationally challenging for realistically sized models and involves approximating the computation, especially in human-in-the-loop settings in which low response times are key for keeping the user engaged in the interaction~\cite{teso2021interactive}.

%
%
\paragraph{Fine-grained active learning.}  In applications with large output spaces, the cost of annotating a full label $\vy$ can be excessive. 
One remedy is to \textit{only ask the supervisor to annotate a (small) subset of output variables} $\vQ^* \subseteq \vY$, selected so to be maximally informative of the full label $\vy$, cf.~\citet{tong2000active,small2010margin}.
This finer-grained form of interaction requires careful handling of the uncertainty associated to \textit{partial} predictions.

A principled solution is to select $\vQ^*$ such that the uncertainty over the whole label set $\vY$ is maximally reduced by solving:
\begin{align}
    \vQ^* \defeq \argmax_{\vQ \subseteq \vY}
        & \quad \bbE_{\vq} \left[ U_\theta(\vY \mid \vx) - U_\theta(\vY \mid \vq, \vx) \right]
        \label{eq:select-sublabels}
    \\
    \text{s.t.}
        & \quad a(\vQ, \vx) \le a_{\max}
\end{align}
where  $a(\vQ,\vx)$ of $\vQ$ denotes the cost of annotating $\vq$ and $a_{\max}$ a limit to it.  E.g., for uniform cost over all concepts in $\vY$ the constraint simplifies to $|\vQ| \le a_{\max}$.
Since $\vx$ is constant, the above objective can be simplified to:
\begin{align}
    \argmin_{\vQ \subseteq \vY} \; \bbE_{\vq} \left[ U_\theta(\vY \mid \vq, \vx) \right] \quad \text{s.t.}
         \ \ a(\vQ, \vx) \le a_{\max}
         \label{eq:select-sublabels-max}
\end{align}

If we take $U_\theta$ to be the margin, then the objective involves evaluating and maximizing the conditional margin $M_\theta(\vY\mid\vq,\vx) = \max_{\vy} p_\theta(\vy\mid\vq,\vx)$.  
For the Shannon entropy $H_\theta$, and letting $\vM \defeq \vY \setminus \vQ$, the objective becomes:
\begin{align}
    \bbE_{\vq} \left[ H_\theta(\vY \mid \vq, \vx) \right]
        & = \bbE_{\vq} \left[ H_\theta(\vM \mid \vq, \vx) \right]
    \\
         = H_\theta(\vM \mid \vQ, \vx)
    &
         = H_\theta(\vY\mid\vx) - H_\theta(\vQ\mid\vx)
        \label{eq:maximize-conditional-entropy}
\end{align}
where last step follows from the chain rule of entropies.
Since $H_\theta(\vY\mid\vx)$ is constant w.r.t. $\vQ$, this shows that Eq.~\ref{eq:select-sublabels} is equivalent to maximizing the conditional entropy of $\vQ$ given $\vX = \vx$, namely:
\[
    \textstyle
    H_\theta(\vQ\mid\vx)  = - \sum_{\vq} p_\theta(\vq \mid \vx) \log p_\theta( \vq \mid \vx).
    \label{eq:conditional-entropy}
\]
%
%
Regardless, this formulation is appealing because the conditional entropy is submodular~\cite{fujishige2005submodular}, and therefore sports approximation guarantees~\cite{nemhauser1978analysis,sviridenko2004note} via iterative greedy maximization.
However, this still requires computing the entropy reliably at each iteration for different subsets of labels, which may be intractable in general~\citep{krause2005near} and lead to arbitrarily suboptimal results if the errors of approximating the entropy cannot be safely bounded.

\paragraph{Skeptical learning.}In skeptical learning, the machine observes fully annotated examples $(\vx_1, \noisyvy_1), (\vx_2, \noisyvy_2), \ldots,$ but the labels $\noisyvy_t$ are \emph{noisy} -- due, for instance, to inattention on the annotator's part -- and the machine is allowed to ask the supervisor to double-check and relabel them~\cite{zeni2019fixing,bontempelli2020learning}.
%

Existing skeptical approaches identify suspicious examples by comparing how much the machine trusts the user's annotation $\noisyvy_t$ versus how much it trusts its own prediction $\predvy_t = \argmax_{\vy} p_\theta(\vy\mid\vx_t)$.  Most commonly, the model's \emph{suspiciusness} is defined as the difference in likelihood between the model's guess and the user's annotation, namely:
\[
    S_\theta(\vx_t) \defeq \max_{\vy} \ p_\theta(\vy\mid\vx_t) - p_\theta(\noisyy_t\mid\vx_t).
    \label{eq:suspiciousness}
\]
This measure has been adapted to different classes of models, including deep neural networks~\cite{teso2021interactive} and Gaussian Processes~\cite{bontempelli2020learning}.

\section{Designing \methods: A Property-driven Approach for Guaranteed Reliable Inference}

In this section, we study how we can build a class of models that guarantee to reliably compute the measures of uncertainty we just introduced in the context of active and skeptical learning.  We will focus on models that support exact computation---hence delivering the highest form of reliability---of these queries and that can do so tractably, that is, in time polynomial in their sizes.\footnote{For many of these queries the time will be \textit{linear} in the model size.} 

We carry out our analysis within the framework of probabilistic circuits (PCs)~\citep{vergari2020probabilistic,choi2020pc}, expressive models encoded by computational graphs for which tractable computation of queries of interest can be guaranteed as long as the graph satisfies certain structural properties.  
Specifically, we aim at pinpointing which structural properties of circuits can enable or inhibit the exact computation of the uncertainty queries discussed in the previous section~\citep{vergari2021compositional}.
We denote the resulting model class as \methods.

\paragraph{Probabilistic circuits.}  We start with a brief overview of PCs.  
A PC $p$ over variables $\vX$ is a computational graph built out of input distribution, sum, and product units.
Each input distribution unit $n$ represents a base distribution $p_n(\scope(n))$ over some variables $\scope(n) \subseteq \vX$.
%
%
Each sum and product unit $n$ receives inputs from some input units, denoted $\ch(n)$, and computes their weighted sum $\sum_{c\in\ch(n)} w_{c} p_c(\scope(c))$, with $w_{c}\geq 0$, or product $\prod_{c\in\ch(n)} p_c(\scope(c))$, respectively.
The last unit in the graph encodes the joint distribution $p(\vX)$.
The probability of an assignment $p(\vx)$ is readily obtained in a single feedforward evaluation of the PC, and has a computational cost linear in the size of the PC, i.e., the number of edges, denoted as $\mid p\mid$.

%
PCs can be adapted to encode conditional distributions $p(\vY\mid\vx)$  in a number of ways. 
\citet{shen2018conditional} explicitly model the conditionining operations as a circuit. 
However, this can be done tractably only for discrete variables $\vX$ with finite domains.
As it explicitly requires to enumerate all possible conditioning states, this approach hardly scales to numbers of variables beyond tens.
Instead, the amortized approach introduced in \citet{shao2020conditional,shao2022conditional} requires only a single circuit structure over $\vY$, that is shared across all possible input states $\vx$. 
Conditional dependence is realized by having the parameters of such a circuit being modeled by an external module, e.g., a regressor that takes $\vX$ as input.

\paragraph{Structural properties of PCs.}
The key feature of PCs is that many complex functions of the distribution that they encode can be computed in polytime as long as the circuit satisfies certain structural properties over their scope or support.
In the next sections, we will make use of the following structural properties: \emph{smoothness}, \emph{decomposability}, \emph{compatibility}, and \emph{determinism}.
While we refer the reader to~\cite{vergari2021compositional,choi2020pc,darwiche2002knowledge} for an in-depth treatment, the next propositions restate some important results linking these properties to the tractability of core operations that will appear in the next sections: \textit{marginalization} and \textit{maximization} 
of circuits encoding a conditional distribution $p(\vY\mid\vx)$.
We start with marginalization, which is a key operation when computing several uncertainty measures like entropies, as we would need to sum over all possible label configurations $\vy$, cf. Eq.~\ref{eq:entropy}.  Marginalization is tractable for all smooth and decomposable circuits.

\begin{prop}[Tractable marginalization, \citep{shao2020conditional,choi2020pc}]
\label{prop:trac-integration}
Let $p(\vZ\mid\vx)$ be a circuit that is smooth and decomposable over $\vZ$ with input functions that can be tractably marginalized out. Then for any variables $\vY\subseteq\vZ$ and their assignment $\vy$, the marginalization $\sum_{\noisyvy\in\val(\noisyvY)}p(\vy,\noisyvy\mid\vx)$ can be computed exactly in time linear in the size of $p$, where $\noisyvY$ denotes $\vZ\setminus{\vY}$.
\end{prop}

Maximization, also known as MAP inference~\citep{koller2009probabilistic} or most probable explanation~\citep{darwiche2009modeling}, is pivotal in the basic classification setting we are in. The key ingredient for it is determinism.

\begin{prop}[Tractable maximization, \citep{darwiche2009modeling,choi2020pc}]
\label{prop:trac-MAP}
Let $p(\vY\mid\vx)$ be a circuit that is deterministic and decomposable over $\vY$ with input functions that can be tractably maximized. 
Then the maximization problem $\argmax_{\vy}p(\vy\mid\vx)$
can be computed exactly in time linear in the size of $p$.
\end{prop}

\paragraph{Expressive and fast randomized circuits.}  Ensuring that a PC $p_{\theta}(\vY\mid\vX)$ that satisfies the above structural properties to be both compact and feature high-capacity is challenging.
For \methods, we tackle this challenge by realizing an amortized conditional circuit as in~\citet{shao2020conditional} and using deep neural networks for its parameters. 
Conditional circuits in \methods are composed of two elements: a distribution $p_{g(\vX)}(\vY)$ implemented as a PC and a deep gating function $g(\vX)$ that, given an input $\vx$, outputs the parameters $\theta$ of the circuit.  
Intuitively, this means that a \method associates a different conditional distribution over the labels to each $\vx$, for improved flexibility and compactness.
One benefit of this approach is that the neural network can be obtained by replacing the top softmax/sigmoid layer of a \textit{pretrained} network with a different learned gating function.
Another benefit is that the whole model, including $g$, can be trained (or just fine-tuned) end-to-end.

To avoid the need of learning the computational graph of $p$, we propose to adopt a randomly structured circuit -- modified to satisfy the structural properties we need.
This reduces learning a \method to learning its parameters, which can be done by fast gradient-based optimizers (like Adam~\cite{kingma2015adam}).
This construction is convenient also because, by exploiting compact tensorized representations of PCs, it is possible to implement learning seamlessly and scalably using GPU-accelerated libraries such as pytorch~\citep{peharz2020random,peharz2020einsum}.

Now, this randomized construction enforces smoothness and decomposability, but it does violate determinism as well as compatibility with itself (a property also called structured-decomposability~\citep{pipatsrisawat2008new}), two properties that are needed to compute uncertainty queries tractably.
To enforce determinism, we propose to apply a randomized construction as in~\citet{shih2020probabilistic}, where one simultaneously conditions on random sets of variables. 
To guarantee structured-decomposability, on the other hand, we need to constraint the way the random computational graph is built.
Starting from the algorithm in~\citet{peharz2020random}, we can systematically realize a ``template'' for decomposing scopes, also called a \textit{region graph}~\citep{poon2011sum} that
ensures a structured-decomposable circuit as output.

Lastly, \methods model learned in this way can naturally be made consistent with respect to given background knowledge \BK encoded as logical formulas over the input and output variables.
This can be achieved by compiling the logical formulas into a compact circuit, for instance, an arithmetic or logical circuit~\citep{darwiche2002knowledge,choi2013compiling}, that shares the same structural properties as circuit of the \method model.
The remaining step is to ensure that the support of $p_{g(\vX)}(\vY)$ avoids all infeasible output configurations, it is sufficient to multiply it with the circuit encoding \BK, an operation that becomes tractable whenever the two circuits are compatible~\citep{vergari2021compositional}.

\section{Active Learning with \methods}

We start by showing how the commonly used measures of uncertainty $U_{\theta}(\vx)$ can be tractably computed with \methods.
We begin with Shannon entropy.

\begin{prop}[Tractable uncertainty with \methods]
\label{prop:trac-ent}
Let $p_{\theta}(\vY\mid\vX)$ be representable as a \method circuit, then computing its Shannon Entropy over $\vY$ for an input configuration $\vx$, as defined in Eq.~\ref{eq:entropy}, or its margin, as defined in Eq.~\ref{eq:margin}, can be done in time linear in $\mid p\mid$.
\begin{proof}
The tractability of the margin follows directly from Proposition~\ref{prop:trac-MAP}.
As for the entropy, notice that for any input $\vx$ the conditional circuit $p_{g(\vX)}(\vY)$ is equivalent to an \textit{unconditional} circuit $p_\theta(\vY)$, with $\theta = g(\vx)$.  For \methods, this circuit is by construction smooth, decomposable, and deterministic.
Therefore, they support the tractable computation of the Shannon entropy as proved in \citet{vergari2021compositional}.
\end{proof}
\end{prop}

Let us now consider fine-grained active learning.  In this case, computing the Shannon entropy or the margin, solving Eq.~\ref{eq:select-sublabels-max} -- either via combinatorial search or greedy maximization -- involves marginalizing over an arbitrary set of variables in each step.  
Unfortunately, even \methods circuit can be intractable after marginalizing out some variables, unless we compromise on expressiveness, as shown by the following proposition.

\begin{prop}
Let $p_{\theta}(\vY\mid\vX)$ be representable as a \method circuit. Then, computing the conditional Shannon entropy $H_{\theta}(\vQ\mid\vx)$ or conditional margin $M_{\theta}(\vQ\mid\vx)$ for all possible $\vQ\subset\vY$ is intractable in general, unless $p_{\theta}$ encodes a fully-factorized distribution.
\begin{proof}
Again, this follows from reasoning over the unconditional circuit $p_{\theta}(\vY)$ with $\theta = g(\vX)$.  Recall that for \methods $p_\theta$ is deterministic.  For a fixed $\vQ\subset\vY$, we wish to compute $H_{\theta}(\vQ\mid\vx)$ and $M_{\theta}(\vQ\mid\vx)$.  This can be done tractably 
if the circuit obtained after marginalizing out $\vY\setminus\vQ$ is itself deterministic.
{This property is called \textit{marginal determinism} in~\citet{choi2020pc}.}.  Furthermore, this must hold all possible subsets $\vQ$.  
This restrictive condition can be trivially satisfied if $p_{\theta}$ can be represented as a fully-factorized distribution~\cite[Section~8.4]{choi2020pc}.
\end{proof}
\end{prop}

The proof of the above proposition suggests that:
i)~the computation of the conditional Shannon entropy and the conditional margin face the same tractability challenges, as both require marginal determinism in \methods;
and
ii)~tractable and exact computation of uncertainty in the context of fine-grained active learning can be achieved if we manage to replace these uncertainty measures with an alternative that does \textit{not} require 
marginal determinism to be tractably computed.

Point (i) implies that routines commonly used to approximate marginal MAP queries can be exploited to approximate the marginal Shannon entropy. 
These include search algorithms and bound propagation schemes~\citep{xue2016solving,cheng2012approximating,maua2012anytime,choi2021solving} or sampling~\citep{krause2005near}.
The price to pay is that the resulting computation is no longer exact.  Even worse, the uncertainty would be approximated \textit{in each step} of the search, leading to the accumulation of approximation errors and leading to query subsets that are arbitrarily far away from the optimum.

%
Point (ii), however, hints at an alternative strategy that does not involve approximating the uncertainty and that can still be tractably computed with \methods.
We propose to quantify uncertainty using the \emph{\Renyi entropy}, a generalization of Shannon entropy defined as follows:
\[
    R^\alpha_\theta(\vY\mid\vx) = \frac{1}{1 - \alpha} \log \left( \sum_{\vy \in \{0, 1\}^c} p_\theta(\vy\mid\vx)^\alpha \right)
    \label{eq:renyi}
\]
for all $\alpha > 0$, $\alpha \ne 1$.
In the following, we let the conditional \Renyi entropy $R^\alpha_\theta(\vQ\mid\vx)$ for a subset of labels $\vQ\subseteq\vY$ be the quantity obtained by applying Eq.~\ref{eq:renyi} to the conditional distribution $p_\theta(\vQ\mid\vx)$.

The \Renyi entropy is a perfectly valid measure of uncertainty and it converges to the Shannon entropy as $\alpha \to 1$.
More interestingly, it acts as a lower bound of $H_{\theta}(\vx)$ for all $\alpha > 1$, i.e., $R^\alpha_\theta(\vQ\mid\vx) \le H_\theta(\vQ\mid\vx)$.  This suggests that, if we can compute it exactly and tractably for all possible label subsets, we can safely maximise it in place of the Shannon entropy in Eq.~\ref{eq:select-sublabels-max}. The next proposition shows that for \methods this is indeed possible.

\begin{prop}[Tractable \Renyi Entropy of \methods]
\label{prop:trac-ent}
Let $p_{\theta}(\vY\mid\vX)$ be representable as a \method circuit, then computing the its \Renyi Entropy over $\vQ$ for an arbitrary label subset $\vQ\subseteq\vY$ and for $\alpha\in\mathbb{N}, \alpha>1$ and an input configuration $\vx$ can be done in time $\mathcal{O}(\mid p\mid^{\alpha})$.
\begin{proof}
The proof follows from turning $p_{g(\vX)}(\vY)$ into an uncoditional circuit $p_\theta$, which by construction of \methods is structured-decomposable and smooth.  The idea is to first marginalize out the labels $\vY\setminus\vQ$, which is doable in time linear in the size of  $p_\theta$ (Proposition~\ref{prop:trac-integration}), and then compute the unconditional \Renyi entropy by computing the $\alpha$ power of the distribution in  $\mathcal{O}(\mid p\mid^{\alpha})$, cf.~\citet{vergari2021compositional}.
\end{proof}
\end{prop}

The $\alpha$-power\footnote{$\alpha=2$ would suffice and keep the computational cost under control.} circuit of $p_{g(\vX)}(\vY)$ can be materialized (as a tensorized circuit) just once, and then reused for all subsequent computations.  This makes it possible to compute the \Renyi entropy for different choices of $\vQ$ efficiently and exactly.  In turn, this makes it straightforward to find a high-quality uncertain label subset using a branch-and-bound procedure leveraging the tensorized circuit to reliably evaluate the uncertainty associated to each partial configuration.

\section{Skeptical Learning with \methods}

Skeptical learning, a realistic setting in which the machine monitors for incoming examples $(\vx_t, \noisyvy_t)$ that may be mislabeled~\cite{zeni2019fixing}.  
As we mentioned, the machine suspiciousness can be modeled by tracking the margin:
\[
    S_\theta(\vx_t) \defeq \max_{\vy} \ p_\theta(\vy\mid\vx_t) - p_\theta(\noisyy_t\mid\vx_t)
\]
between the user's annotation $\noisyvy_t$ and the machine's prediction $\predvy_t = \argmax_{\vy} p_\theta(\vy\mid\vx_t)$.  It is easy to see that \methods also enable tractable computation of the model's suspiciousness.  This follows directly from Proposition~\ref{prop:trac-ent} by noticing that $p_\theta(\noisyy_t\mid\vx_t)$ is constant with respect to $\vy$.  Naturally, as for active learning, this result holds even under hard constraints between the output labels, a feature that is not supported by any other skeptical learning approach (see the Related Work below for a more detailed breakdown)

\section{Related Work}

\paragraph{PC learning.} Starting from \cite{lowd2005naive}, learning the structure and parameters of tractable models that can be represented as PCs has been an active research field.
These include arithmetic circuits~\citep{darwiche2009modeling}, probabilistic sentential decision diagrams~\citep{kisa2014probabilistic}, sum-product networks~\citep{poon2011sum}, and cutset networks~\citep{rahman2014cutset}.
For a survey, see~\citep{vergari2020probabilistic}.
The vast majority of these structure learning algorithms, however, are non-differentiable and hence cannot be seamlessly integrated with deep neural networks.
Randomized approaches \citep{di2017fast,di2015learning,peharz2020random,peharz2020einsum,di2021random,DBLP:conf/pgm/VentolaS0K20,DBLP:conf/aiia/MauroEVV17} alleviate this issue.
An orthogonal direction is to employ ensembles to boost model accuracy \citet{vergari2015simplifying,dang2022strudel,di2017fast,rahman2016learning} and it is potentially applicable to \methods.

\paragraph{Active learning.}  Most work on active learning from sequential data focuses on simple shallow classifiers (e.g., linear separators) and neglects the issues of learning in the wild~\cite{hoi2021online}.  Active learning with partial labels has been studied in the context of Bayesian networks~\cite{tong2000active}, max-margin predictors~\cite{roth2006margin,small2010margin}, and other models~\cite{sun2015active,mo2016learning,liu2017active,khodabandeh2017active,hu2018active,behpour2019active,ning2019partial,nakano2020active}.
Closest to our work, 
\citet{platanios2017active} also develop selection heuristics for picking informative example/sub-label pairs based on entropy reduction.
\citet{luo2013latent}, instead, look at pool-based active learning for structured-output prediction with latent variables and propose to acquire sub-labels with high \emph{marginal} entropy.
These approaches do not sport reliable subset computation, however.
%
%
 %
\citet{krishnamurthy2017active} look at online learning for cost-sensitive multi-class classification problems and propose a theoretically well-founded algorithm, which however does not translate to high-capacity models.
None of the above approaches considers sequential settings, nor provide a unified framework to design models supporting the tractable computation of the uncertainty queries of interest. 

\paragraph{Skeptical learning.}  Existing approaches to skeptical learning~\cite{zeni2019fixing,bontempelli2020learning}, address human-in-the-loop learning tasks with noisy examples.  Another close line of work is learning from weak annotators, which consider a similar setup but in a pool-based setting~\cite{urner2012learning,kremer2018robust}.  In contrast to \methods, these approaches are either restricted to multi-class classification or offer no support for representation learning.  For instance, \citet{zeni2019fixing} tackle hierarchical classification, but they do so using a custom structured-output classifier based on random forests that offers rather sub-optimal uncertainty estimates~\cite{bontempelli2020learning}.  On the other hand, \citet{teso2021interactive} combine skeptical learning with explanation-based interaction in the context of deep neural networks~\cite{schramowski2020making}, but are unconcerned with structured output spaces.


\section{Conclusion and Outlook}

We have introduced \methods, a novel class of PCs designed specifically for tasks involving interaction with (unreliable) human agents.
Our key contribution is identifying structural properties of probabilistic circuits that ensure reliable computation of uncertainty in the context of active and skeptical learning.
These include the margin and several entropy formulations.
Then, within this framework, we have shown how \methods can tackle the reliable computation of query label subsets for fine-grained active learning by relying on the \Renyi entropy.
We plan to evaluate \methods on real-world active learning benchmarks for deep learning~\citep{gal2017deep} as well as SOP tasks such as  hierarchical multi-label classification~\citep{giunchiglia2020coherent}.

\section*{Acknowledgements}
The research of ST was partially supported by TAILOR, a project funded by EU Horizon 2020 research and innovation programme under GA No 952215.

\bibliography{imlw}
\end{document}